\documentclass[11pt,a4paper]{article}
\usepackage[hyperref]{acl2021}
\usepackage{times}
\usepackage{latexsym}

\usepackage{microtype}
\usepackage{enumitem}
\usepackage{booktabs}
\usepackage{amsfonts}
\usepackage{eucal}
\usepackage{graphicx}
\usepackage{color}
\usepackage{caption}
\usepackage{subcaption}
\usepackage{multirow}

\aclfinalcopy % Uncomment this line for the final submission
\newcommand{\modelname}{Lawformer}

\title{Lawformer: A Pre-trained Language Model for \\ Chinese Legal Long Documents}

\author{Chaojun Xiao\textsuperscript{\rm 1}$^\ast$,
Xueyu Hu\textsuperscript{\rm 2}\thanks{~~Indicates equal contribution.}~,
Zhiyuan Liu\textsuperscript{\rm 1}\thanks{~~Corresponding author.}~,
Cunchao Tu\textsuperscript{\rm 3},
\textbf{Maosong Sun\textsuperscript{\rm 1} }\\
\textsuperscript{\rm 1}Department of Computer Science and Technology\\
Institute for Artificial Intelligence, Tsinghua University, Beijing, China\\
Beijing National Research Center for Information Science and Technology, China \\
\textsuperscript{\rm 2} Beihang University, Beijing, China \\
\textsuperscript{\rm 3} Beijing Powerlaw Intelligent Technology Co., Ltd., China \\
{\tt xcjthu@gmail.com, huxueyu@buaa.edu.cn} \\
{\tt tucunchao@gmail.com \{liuzy,sms\}@tsinghua.edu.cn}
\\}

\date{}

\begin{document}
\maketitle
\begin{abstract}
Legal artificial intelligence (LegalAI) aims to benefit legal systems with the technology of artificial intelligence, especially natural language processing (NLP). Recently, inspired by the success of pre-trained language models (PLMs) in the generic domain, many LegalAI researchers devote their effort to apply PLMs to legal tasks. However, utilizing PLMs to address legal tasks is still challenging, as the legal documents usually consist of thousands of tokens, which is far longer than the length that mainstream PLMs can process. In this paper, we release the Longformer-based pre-trained language model, named as \modelname{}, for Chinese legal long documents understanding. We evaluate \modelname{} on a variety of LegalAI tasks, including judgment prediction, similar case retrieval, legal reading comprehension, and legal question answering. The experimental results demonstrate that our model can achieve promising improvement on tasks with long documents as inputs.
The code and parameters are available at \url{https://github.com/thunlp/LegalPLMs}.
\end{abstract}

\section{Introduction}

Legal artificial intelligence (LegalAI) focuses on applying methods of artificial intelligence to benefit legal tasks~\cite{zhong2020does}, which can help improve the work efficiency of legal practitioners and provide timely aid for those who are not familiar with legal knowledge. Thus, LegalAI has received great attention from both natural language processing (NLP) researchers and legal professionals~\cite{zhong2018legal,zhong2020jec,wu2020biased,chalkidis2020legal,hendrycks2021cuad}.

In recent years, pre-trained language models~(PLMs)~\cite{peters2018deep,devlin2019bert,liu2019roberta,raffel2020exploring,Brown2020language} have proven effective in capturing rich language knowledge from large-scale unlabelled corpora and achieved promising performance improvement on various downstream tasks. Inspired by the great success of PLMs in the generic domain, considerable efforts have been devoted to employing powerful PLMs to promote the development of LegalAI~\cite{shaghaghian2020customizing,shao2020bert,chalkidis2020legal}.

Some researchers attempt to transfer the contextualized language models pre-trained on the generic domain, such as Wikipedia, Children’s Books, to tasks in the legal domain~\cite{shaghaghian2020customizing,zhong2020jec,shao2020bert,elwany2019bert}. Besides, some works conduct continued pre-training on legal documents to bridge the gap between the generic domain and the legal domain~\cite{zhong2019openclap,chalkidis2020legal}.

\begin{figure}[t]
    \centering
    \includegraphics[width=0.9\columnwidth]{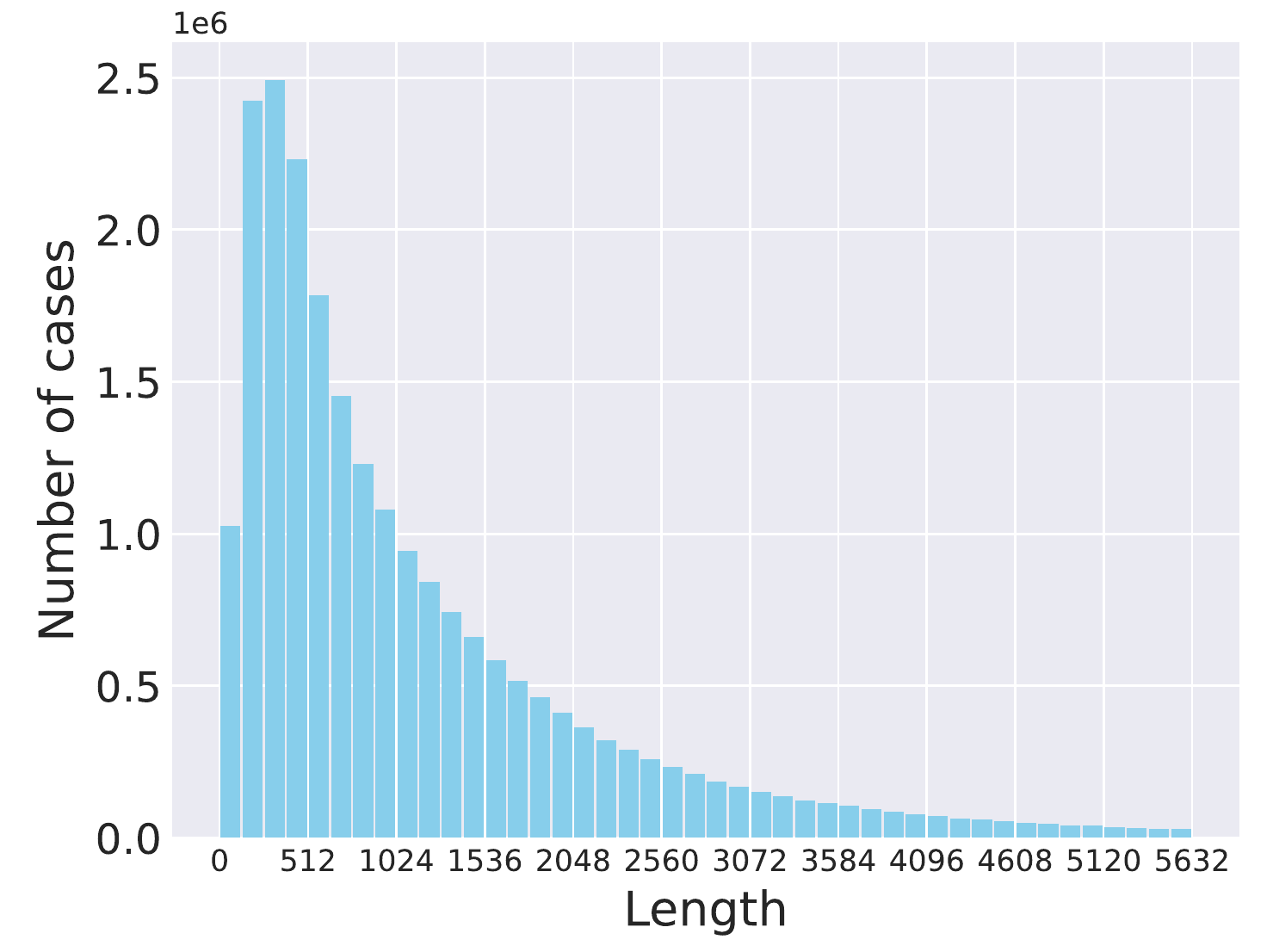}
    \caption{The length distribution of the fact description in criminal and civil cases. The criminal and civil cases consist of $1260.2$ tokens on average. The length of $64.12$\% cases is over $512$.}
    \label{fig:intro_length}
\end{figure}

However, most of these works adopt the full self-attention mechanism to encode the documents and cannot process long documents due to the high computational complexity. As shown in Figure~\ref{fig:intro_length}, the average length of the criminal cases and civil cases is $1260.2$, which is far longer than the maximum length that mainstream PLMs (BERT, RoBERTa, etc.) can handle. With the limited capacity to process long sequences, these PLMs cannot achieve satisfactory performance in representing the legal documents~\cite{zhong2020does,shao2020bert}. Hence, how to utilize PLMs to process legal long sequences needs more exploration.

In this work, we release \modelname{}, which is pre-trained on large-scale Chinese legal long case documents. \modelname{} is a Longformer-based~\cite{beltagy2020longformer} language model, which can encode documents with thousands of tokens. Instead of employing the standard full self-attention, we combine the local sliding window attention and the global task motivated full attention to capture the long-distance dependency.  To the best of our knowledge, \modelname{} is the first legal pre-trained language model, which can process the legal documents with thousands of tokens.

Besides, we evaluate \modelname{} on a collection of typical legal tasks including legal judgment prediction, similar case retrieval, legal reading comprehension, and legal question answering.
Solving these tasks requires the model to understand domain knowledge and concepts in legal texts, and be able to analyze complicated case scenarios and legal provisions. 

Notably, the data distribution of existing datasets for legal judgment prediction is quite different from the real-world data distribution. And these datasets only contain criminal cases and omit civil cases. Therefore, we construct new datasets from scratch for legal judgment prediction tasks, which consist of hundreds of thousands of criminal cases and civil cases. As for the other tasks, we rely on the pre-existing datasets. Experimental results on these various tasks demonstrate that the proposed \modelname{} can achieve strong performance in legal documents understanding.

The main contributions of this paper are summarized as follows:
\begin{itemize}[leftmargin=*,itemsep=0pt,topsep=0pt]
    \item We release a Chinese legal pre-trained language model, which can process documents with thousands of tokens, named as \modelname{}. To the best of our knowledge, \modelname{} is the first pre-trained language model for legal long documents.
    \item We evaluate \modelname{} for legal documents understanding, with high coverage of existing typical LegalAI tasks. In this benchmark, we propose new legal judgment prediction datasets for both criminal and civil cases.
    \item Extensive experiments demonstrate the proposed \modelname{} can achieve strong performance on various LegalAI tasks that require the models are able to process the long documents. In terms of the tasks with short inputs, \modelname{} can also achieve comparable results with RoBERTa~\cite{liu2019roberta} pre-trained on the legal corpora.
\end{itemize}

\section{Related Work}
\subsection{Legal Artificial Intelligence}
Legal artificial intelligence aims to profit the tasks in the legal domain~\cite{zhong2020does}. Due to the amount of textual legal resources, LegalAI has drawn great attention from NLP researchers in recent years~\cite{luo2017learning,ye2018interpretable,duan2019legal,shao2020bert,wu2020biased}. Early works attempt to analyze legal documents with hand-crafted features and statistical methods~\cite{kort1957predicting,nagel1963applying,segal1984predicting}. With the development of deep learning, many efforts have been devoted to solving various legal tasks, such as legal charge prediction~\cite{luo2017learning,zhong2018legal,xiao2018cail2018,chalkidis2019neural,yang2019legal}, relevant law article retrieval~\cite{chen2013text,raghav2016analyzing}, court view generation~\cite{ye2018interpretable,wu2020biased}, reading comprehension~\cite{duan2019cjrc}, question answering~\cite{zhong2020jec,kien2020answering}, and case retrieval~\cite{raghav2016analyzing,shao2020bert}. Besides, inspired by the great success of PLMs in the generic domain, there are also some researchers conducting pre-training on legal corpora~\cite{zhong2019openclap,chalkidis2020legal}. However, these models adopt the BERT as the basic encoder, which cannot process documents longer than $512$. To the best of our knowledge, \modelname{} is the first pre-trained language model for legal long documents.

\subsection{Pre-trained Language Model}
Pre-trained language models (PLMs), which are trained on amounts of unlabelled corpora, are able to benefit a variety of downstream NLP tasks~\cite{devlin2019bert,liu2019roberta,Brown2020language}. Then we will introduce previous works related to ours from the following two aspects: domain-adaptive pre-training and long-document pre-training.

\begin{figure*}[ht]
    \centering
    \includegraphics[width=0.75\textwidth]{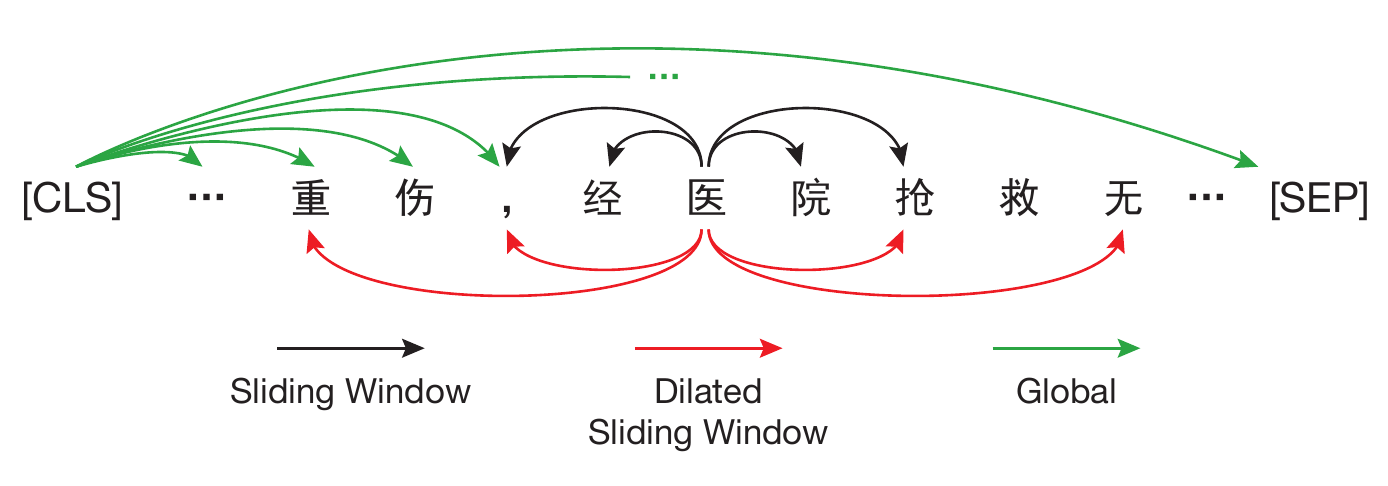}
    \caption{An example of the combination of the three types of attention mechanism in \modelname{}. The size of the sliding window attention is $2$. The size of the dilated sliding window attention is $2$ and the gap is $1$. The token, \texttt{[CLS]}, is selected to perform the global attention.}
    \label{fig:attention}
\end{figure*}

\textbf{Domain-Adaptive Pre-training.} 
Many researchers attempt to achieve performance gain by domain-adaptive pre-training on various domains, including the biomedical domain~\cite{Lee2020BioBERT}, the clinical domain~\cite{alsentzer2019publicly}, the scientific domain~\cite{beltagy2019scibert}, and the legal domain~\cite{zhong2019openclap,chalkidis2020legal}. These works further pre-train the BERT~\cite{devlin2019bert} on the specific domain texts and have shown that continued pre-training on the target domain corpora can consistently achieve performance improvement~\cite{gururangan2020don}.

\textbf{Long-Document Pre-training.} Due to the high computational complexity of the full self-attention mechanism, the traditional transformer-based PLMs are limited in processing long documents~\cite{beltagy2020longformer}. Some works propose to utilize the left-to-right auto-regressive objective to pre-train the language models~\cite{dai2019transformer,sukhbaatar2019adaptive}. And some works attempt to reduce the computational complexity with sliding window based self-attention~\cite{beltagy2020longformer,qiu2020blockwise,zaheer2020big}. In this work, we adopt the widely-used Longformer~\cite{beltagy2020longformer} as our basic encoder, which combines the sliding window attention and the global full attention to process the documents. 

\section{Our Approach}

\subsection{\modelname{}}
Our current model utilizes Longformer~\cite{beltagy2020longformer} as our basic encoder. Instead of utilizing the full self-attention mechanism, Longformer combines the sliding window attention, dilated sliding window attention, and the global attention mechanism to encode the long sequence. We will introduce the three types of attention patterns in the section. Figure~\ref{fig:attention} gives an example of the combination of three types of attention mechanisms.

\textbf{Sliding Window Attention.} In this attention pattern, we only compute the attention scores between the surrounding tokens. Specifically, given the size of the sliding window $\mathit{w}$, each token only attends to the $\frac{1}{2}\mathit{w}$ tokens on each side. While the tokens will only aggregate information around them in each layer, with the number of layers increases, the global information can also be integrated into the hidden representations of each token.

\textbf{Dilated Sliding Window Attention.} Similar to dilated CNNs~\cite{vanwavenet}, the sliding window attention can be dilated to reach a longer context. In this attention mechanism, each window is not continuous but has a gap $d$ between each attended token. Notably, in the multi-head attention, the gap in different heads can be different, which can promote the model performance.

\textbf{Global Attention.} In some specific tasks, we need some tokens to attend to the whole sequence to obtain sufficient information. For example, in the text classification task, the special token \texttt{[CLS]} should be used to attend to the whole document. In the question answering task, the questions are supposed to attend the whole sequence to generate expressive representations. Therefore, we apply global attention for some pre-selected tokens for task-specific representations. That is, instead of attending to the surrounding tokens, the selected tokens will attend to the whole sequence to generate the hidden representations. Notably, the parameters in the global attention and sliding window attention are different.

With the three types of attention mechanisms, we can process the long sequences with linear complexity.

\begin{table*}[t]
\centering
\small
\begin{tabular}{l ccccc cccc}
\toprule
          & \multicolumn{5}{c}{criminal} & \multicolumn{4}{c}{civil} \\ \cmidrule(lr){2-6} \cmidrule(lr){7-10}
Model     &  Mic@c  &  Mac@c  &  Mic@l  &  Mac@l   &  Dis@t  &  Mic@c  &  Mac@c  &  Mic@l  &  Mac@l   \\ \midrule
BERT      & 94.8  &  68.2 &  81.5 &  52.9  & 1.286 & 80.6  &  47.6 &  61.7 & 31.6 \\
RoBERTa   & 94.7  &  69.3 &  81.1 &  53.5  & 1.291 & 80.0  &  47.2 &  60.2 & 29.9 \\
L-RoBERTa & 94.9  &  70.8 &  81.1 &  53.4  & 1.280 & 80.8  &  49.4 &  61.2 & 31.3 \\ \midrule
Lawformer & \textbf{95.4} & \textbf{72.1}  & \textbf{82.0} & \textbf{54.3}  & \textbf{1.264}
          & \textbf{81.1} & \textbf{50.0} & \textbf{63.0} & \textbf{33.0} \\ \bottomrule
\end{tabular}
\caption{The results on the legal judgment prediction dataset, CAIL-Long. For criminal cases, we evaluate the models on charge prediction task (Mic@c, Mac@c), relevant law prediction task (Mic@l, Mac@l), and term of penalty prediction task (Dis@t). For civil cases, we evalute the models on cause of actions prediction task (Mic@c, Mac@c), and relevant law prediction task (Mic@l, Mac@l).}
\label{tab:ljp}
\end{table*}

\subsection{Data Processing}
We collect tens of millions of case documents published by the Chinese government from China judgment Online\footnote{https://wenshu.court.gov.cn/}. As the downstream tasks are mainly in the areas of criminal and civil cases, we only keep the documents of criminal cases and civil cases. We divide each document into four parts: the information about the parties, the fact description, the court views, and the judgment results. We only keep the documents with the fact description longer than $50$ tokens. After the data processing, the rest of the data are used for pre-training. The detailed statistics is listed in Table~\ref{tab:pretrain_data}.

\begin{table}[h]
\small
\centering
\begin{tabular}{lrrr}
\toprule
               &   \# Doc.  &  Len.  & Size \\ \midrule
criminal cases & 5,428,717  &  962.84   & 17 G   \\
civil cases    & 17,387,874 & 1,353.03  & 67 G  \\ \bottomrule
\end{tabular}
\caption{The statistics of the pre-training data. \# Doc. refers to the number of documents. Len. refers to the average length of the documents. Size refers to the size of the pre-training data.}
\label{tab:pretrain_data}
\end{table}

\subsection{Pre-training Details}
Following the previous work~\cite{beltagy2020longformer}, we pre-train \modelname{} with MLM objective, continuing from the checkpoint, RoBERTa-wwm-ext, released in~\citet{cui2019pre}. 
We set the learning rate as $5 \times 10^{-5}$, the sequence length as $4,096$, and the batch size as $32$. As the length of legal documents is usually smaller than $4,096$, we concatenate different documents together to make full use of the input length. We pre-train \modelname{} for $200,000$ steps, and the first $3,000$ steps are for warm-up. We utilize Adam~\cite{kingma2015adam} to optimize the model. The rest of the hyper-parameters are the same as Longformer. We pre-train \modelname{} with $8 \times 32$G NVIDIA V100 GPUs.

In the fine-tuning stage, we select different tokens to conduct the global attention mechanism. For the classification task, we select the token, \texttt{[CLS]}, to perform the global attention. And for the reading comprehension task and the question answering task, we perform the global attention on the whole questions. For the specific details of each task, please refer to the next section.

\section{Experiments}

\subsection{Baseline Models}
To verify the effectiveness of the proposed model, we compare \modelname{} with following competitive baseline models:
\begin{itemize}[leftmargin=*,itemsep=0pt,topsep=0pt]
    \item BERT~\cite{devlin2019bert}: we simply fine-tune the published checkpoint, BERT-base-chinese, which is pre-trained on Chinese wikipedia documents\footnote{https://zh.wikipedia.org/}, on the following downstream datasets.
    \item RoBERTa-wwm-ext (RoBERTa)~\cite{cui2019pre}: it is pre-trained with the whole word masking strategy, in which the tokens that belong to the same word will be masked simultaneously. Notably, \modelname{} is pre-trained continuously from the RoBERTa.
    \item Legal RoBERTa (L-RoBERTa): we pre-train a RoBERTa~\cite{liu2019roberta} on the same legal corpus, continuing from the released RoBERTa-wwm-ext checkpoint.
\end{itemize}

As these baseline models can only process documents with less than $512$ tokens, we only truncate the documents to $512$ tokens for these models.

\begin{table*}[t]
\centering
\small
\begin{tabular}{lccccccccc}
\toprule
Model     &  P@5  & P@10  &  P@20 &  P@30 & NDCG@5 & NDCG@10 & NDCG@20 & NDCG@30 & MAP \\ \midrule
BERT      & 44.27 & 41.83 & 36.73 & 33.49 &  78.18 &  80.06  &  84.43  &  91.46  & 50.65 \\
RoBERTa   & 45.93 & 41.71 & 36.53 & 33.40 &  79.93 &  80.57  &  84.99  &  91.82  & 50.77 \\
L-RoBERTa & 45.75 & 42.85 & 37.79 & 33.58 &  78.90 &  81.01  &  85.26  &  91.70  & 50.17 \\ \midrule
Lawformer & \textbf{51.91} & \textbf{46.44} & \textbf{37.95} & \textbf{33.99} & \textbf{83.11} & \textbf{84.05} & \textbf{87.06} & \textbf{93.22} & \textbf{57.36} \\ \bottomrule
\end{tabular}
\caption{The results on the legal case retrieval dataset, LeCaRD. The dataset contains long cases with thousands of tokens as candidates.}
\label{tab:lecard}
\end{table*}

\subsection{Legal judgment Prediction}

\textbf{Dataset Construction:} Legal judgment prediction aims to predict the judgment results given the fact description. It is a critical application in the LegalAI field and has received great attention recently. To facilitate the development of this task, \citet{xiao2018cail2018} have proposed a large-scale criminal judgment prediction dataset, CAIL2018. However, the average case length of CAIL2018 is much shorter than that of real-world cases. Besides, CAIL2018 only focuses on criminal cases and omits civil cases. In this paper, we propose a new judgment prediction dataset, CAIL-Long, which contains both civil and criminal cases with the same length distribution as in the real world.

CAIL-Long consists of $1,129,053$ criminal cases and $1,099,605$ civil cases. For both criminal and civil cases, we take the fact description as inputs and extract the judgment annotations with regular expressions. Specifically, each criminal case is annotated with the charges, the relevant laws, and the term of penalty. Each civil case is annotated with the causes of actions and the relevant laws. The detailed statistics of the dataset are shown in Table~\ref{tab:cail-long}.

\begin{table}[h]
\centering
\small
\resizebox{\columnwidth}{!}{
\begin{tabular}{l rrrrr}
\toprule
          & \# Case & Len. & \# C. & \# Law & prison \\ \midrule
criminal  & 115,849 & 916.57 & 201 & 244 & 0-180  \\
civil     & 113,656 & 1,286.88 & 257 & 330 & --  \\ \bottomrule

\end{tabular}
}
\caption{The statistics of the judgment prediction datset, CAIL-Long. \# Case denotes the number of cases. Len. denotes the average length of the fact description. \# C. denotes the number of charges/cause of actions. \# Law denotes the number of relevant laws. And prison denotes the range of term of penalty (unit: month).}
\label{tab:cail-long}
\end{table}

\smallskip
\noindent
\textbf{Implementation Details:} Following previous work~\cite{zhong2018legal}, we train the models in the multi-task paradigm. For criminal cases, the charge prediction and the relevant law prediction are formalized as multi-label text classification tasks. The term of penalty prediction task is formalized as the regression task. For civil cases, the cause of actions prediction is formalized as a single-label text classification task, and the relevant law prediction is formalized as a multi-label text classification task. The models for criminal and civil cases are trained separately. 

We adopt micro-F1 scores (Mic@\{c,l\}) and macro-F1 scores (Mac@\{c,l\}) as metrics for classification tasks, and the log distance (Dis@t) as metric for the regression task. We set the learning rate as $5 \times 10^{-5}$, and batch size as $32$.

\smallskip
\noindent
\textbf{Result:} We present the results in Table~\ref{tab:ljp}. As shown in the table, \modelname{} achieves the best performance among the four models in both the micro-F1 and macro-F1 scores. The improvement indicates that \modelname{} can accurately capture the key information from the long fact description. Besides, the case numbers of different labels (charges, cause of actions, and laws) are highly imbalanced, and \modelname{} can also achieve significant improvement in macro-F1 scores, which indicates that \modelname{} is able to handle the labels with limited cases. However, the overall results are still unsatisfactory. It still needs more exploration to predict the judgment results more accurately and robustly.

\subsection{Legal Case Retrieval}

\textbf{Dataset:} Legal case retrieval is the specialized information retrieval task in the legal domain, which aims to retrieve similar cases given the query fact description. For this task, we adopt the Legal Case Retrieval Dataset (LeCaRD)\footnote{https://github.com/myx666/LeCaRD} as our benchmark, which contains $107$ query cases and $10,716$ candidate cases. Notably, the length of the candidate cases is extremely long. As shown in the Table~\ref{tab:lecard_res}, the average length of the cases is $6,319.14$, which is a great challenge for the existing models.

\begin{table}[h]
\centering
\small
\begin{tabular}{ccccc}
\toprule
\# Query & \# Cand. & Q-Len. & C-Len. & \# Pair  \\ \midrule
107& 10,716 & 444.55 & 6,319.14 & 1,094  \\ \bottomrule
\end{tabular}
\caption{The statistics of LeCaRD dataset. \# Query and \# Cand. denote the number of query cases and candidate cases. Q-Len. and C-Len. denote the average length of the fact description of the query cases and candidate cases. \# Pair denotes the number of positive query-candidate case pairs.}
\label{tab:lecard_res}
\vspace{-1em}
\end{table}

\smallskip
\noindent
\textbf{Implementation Details:}
We train the models with the binary classification task, which requires the models to judge whether the given candidate case is relevant to the query case. We set the fine-tuning batch size as $32$, the learning rate as $10^{-5}$ for all models. For the baseline models, which can only process sequences with lengths no more than $512$, we set the maximum length of the query and the candidates as $100$ and $409$, respectively. And for the \modelname{}, we set the maximum length of the query and the candidates as $509$ and $3,072$, respectively. All tokens of the query case are selected to perform the global attention mechanism.

Following the previous works, we adopt $5$-fold cross-validation for the dataset. We employ the top-$k$ Normalized Discounted Cumulative Gain (NDCG@$k$), Precision (P@$k$), and Mean Average Precision (MAP) as evaluation metrics. For each model, we report the performance of the checkpoint with the highest average score on P@$10$, NDCG@$10$, and MAP.

\smallskip
\noindent
\textbf{Result:} The results are shown in Table~\ref{tab:lecard}. We can observe that \modelname{} can significantly outperform all baseline models. For example, \modelname{} improves upon baselines by $6.59$ points in terms of mean average precision. The similar case retrieval task requires the models to compare the query case and candidate case in detail. The baseline models even cannot read the complete documents, and thus perform unsatisfactorily. \modelname{} can process sequences with thousands of tokens and achieve promising results. However, the average length of candidate cases in LeCaRD is $6,319.14$, which is also beyond the capacity of \modelname{}. We argue that it still needs further exploration to retrieve the long cases.

\subsection{Legal Reading Comprehension}
\textbf{Dataset:} We utilize the Chinese judicial reading comprehension (CJRC) as our benchmark dataset for legal reading comprehension~\cite{duan2019cjrc}. CJRC is published in the Chinese AI and Law Challenge contest. We adopt the dataset published in 2020 as the benchmark\footnote{https://github.com/china-ai-law-challenge/CAIL2020}. CJRC consists of $9,532$ question-answer pairs with corresponding supporting sentences. There are three types of answers for these questions, including the span of words, yes/no, and unanswerable. The detailed statistics of CJRC are shown in Table~\ref{tab:cjrc}. It is worth mentioning that the average length of the documents in CJRC is only $441.04$.

\begin{table}[h]
\centering
\small
\resizebox{\columnwidth}{!}{
\begin{tabular}{cccccc}
\toprule
\# Doc & Len. & \# Que. & \# S-Que. & \# YN-Que.  & \# U-Que. \\ \midrule
9,532& 441.04 & 9,532 & 6,692 & 1,892 & 948  \\ \bottomrule
\end{tabular}
}
\caption{The statistics of CJRC dataset. \# Doc denotes the number of case documents. Len. denotes the average length of the documents. \# Que., \# S-Que., \# YN-Que., and \# U-Que. denotes the total number of questions, questions with span of words as answers, questions with yes/no answers, unanswerable questions, respectively.}
\label{tab:cjrc}
\vspace{-1em}
\end{table}

\smallskip
\noindent
\textbf{Implementation Details:} We implement the models following the source code provided in~\citet{duan2019cjrc}. We train the models to predict the start positions and end positions. And the supporting sentence prediction is formalized as the binary classification for all the sentences. For the \modelname{}, we select the whole question to perform the global attention. We adopt the exact match score (EM) and F1 score as evaluation metrics.

\smallskip
\noindent
\textbf{Result:} The results are shown in Table~\ref{tab:LRC}. From the results, we can observe that (1) L-RoBERTa and \modelname{}, which are pre-trained on the in-domain corpus, can significantly outperform the BERT and RoBERTa model, which are pre-trained on the out-of-domain corpus. (2) L-RoBERTa can achieve comparable performance with \modelname{}, as the long documents are filtered out when constructing the dataset. We argue that with the improvement of the ability to process long sequences, \modelname{} can achieve better performance in the long document reading comprehension.

\begin{table}[h]
\centering
\small
\resizebox{\columnwidth}{!}{
\begin{tabular}{l cc cc cc}
\toprule
              & \multicolumn{2}{c}{Answer} & \multicolumn{2}{c}{Support} &  \multicolumn{2}{c}{Joint}\\ \cmidrule(lr){2-3} \cmidrule(lr){4-5} \cmidrule(lr){6-7}
Model         &   EM  &  F1   &   EM  &  F1   &   EM  &  F1  \\ \midrule
BERT          & 50.70 & 66.53 & 31.45 & 71.80 & 21.19 & 52.15 \\
RoBERTa       & 53.81 & 68.86 & 34.60 & 73.63 & 23.76 & 55.62 \\
L-RoBERTa     & 54.12 & 69.76 & \textbf{35.73} & \textbf{74.44} & \textbf{24.42} & 56.40 \\ \midrule
Lawformer     & \textbf{55.02} & \textbf{69.98} & 35.15 & 74.28 & 24.18 & \textbf{56.62} \\ \bottomrule
\end{tabular}
}
\caption{The results on the legal reading comprehension task. Answer refers to the question answering task. Support refers to the supporting sentences prediction task. Joint refers to the overall performance.}
\label{tab:LRC}
\end{table}

\subsection{Legal Question Answering}
\textbf{Dataset:} Legal question answering requires the model to understand legal knowledge and answer the given questions. A high-quality legal question answering system can provide an accurate legal consult service. For this task, we select JEC-QA~\cite{zhong2020jec} to evaluate the performance of the proposed model. JEC-QA consists of $28,641$ multiple-choice questions from the Chinese national bar exam, which is quite challenging for existing models~\cite{zhong2020jec}.

\smallskip
\noindent
\textbf{Implementation Details:} We formalize the task as a text classification task. First, we concatenate the questions and candidate choices together to form the inputs of the models. Then a linear layer is applied to compute the matching scores of the candidates. Previous works also take the reading materials retrieved by statistical methods (BM25, TFIDF) as inputs~\cite{zhong2020jec,zhong2020does}. We ignore these reading materials in our experiments, as the retrieval results are unsatisfactory and would harm the performance.
We set the learning rate as $2 \times 10^{-5}$ and the batch size as $32$. The positions are set as $0$  for questions, and $1$ for candidate choices.

\begin{table}[h]
\centering
\small
\begin{tabular}{l cc cc}
\toprule
              & \multicolumn{2}{c}{dev} & \multicolumn{2}{c}{test} \\ \cmidrule(lr){2-3} \cmidrule(lr){4-5}
Model         & single &  all  & single &  all  \\ \midrule
BERT          & 42.78  & 25.77 & 41.23  & 24.50  \\
RoBERTa       & 44.46  & \textbf{28.18} & 43.18  & 27.09  \\
L-RoBERTa     & 45.17  & 27.54 & 43.29 & \textbf{27.81}  \\ \midrule
Lawformer     & \textbf{45.81} & 27.21 & \textbf{45.81} & 27.43  \\ \bottomrule
\end{tabular}
\caption{The results on the legal question answering task. We adopt accuracy as the evaluation metric. Here, single denotes the single-answer questions and all denotes all questions.}
\label{tab:LegalQA}
\end{table}

\smallskip
\noindent
\textbf{Result:} The results are shown in Table~\ref{tab:LegalQA}. As the inputs in the dataset do not require long-distance understanding, L-RoBERTa can achieve comparable results with \modelname{}. However, as the task needs the models to perform complex reasoning~\cite{zhong2020jec}, all the models cannot achieve promising results. Therefore, it remains a great challenge for future works to enhance the models with legal knowledge and the logic reasoning capacity.

\section{Conclusion and Future Work}
In this paper, we pre-train a Longformer-based language model with tens of millions of criminal and civil case documents, which is named as \modelname{}. Then we evaluate \modelname{} on several typical LegalAI tasks, including legal judgment prediction, similar case retrieval, legal reading comprehension, and legal question answering. The results demonstrate that \modelname{} can achieve significant performance improvement on tasks with long sequence inputs.

Though \modelname{} can achieve performance improvement for legal documents understanding, the experimental results also show that the challenges still exist. In the future, we will further explore the legal knowledge augmented pre-training. It is an established fact that enhancing the models with legal knowledge is quite necessary for many LegalAI tasks~\cite{zhong2020does}.

Meanwhile, we will also explore the generative legal pre-trained model. In real-world legal practice, the practitioners need to conduct heavy and redundant paper writing works. A powerful generative legal pre-trained language model can be beneficial for legal professionals to improve work efficiency.

To summarize, we release \modelname{} for legal long document understanding in this paper. In the future, we will attempt to integrate legal knowledge into the legal pre-trained language models, and pre-train generative models for LegalAI tasks.

\bibliographystyle{acl_natbib}
\bibliography{anthology,acl2021}

%\appendix

\end{document}